%% file: emnlp2020.tex
\documentclass[11pt,a4paper]{article}
\usepackage[hyperref]{emnlp2020}
\usepackage{times}
\usepackage{latexsym}
\usepackage{graphicx}
\usepackage{bm}
\usepackage{multirow} 
\usepackage{amsmath}
\usepackage{amssymb}
\usepackage{amsfonts}
\usepackage{booktabs}

\usepackage{microtype}

\aclfinalcopy 
\def\aclpaperid{158} 

\title{Double Graph Based Reasoning for Document-level Relation Extraction}

\author{
  Shuang Zeng$^{1,2}$\footnotemark[1], 
  Runxin Xu$^{1}$\footnotemark[1], 
  Baobao Chang$^{1,2}$\footnotemark[2] \and 
  Lei Li$^{3}$\\
  $^{1}$Key Laboratory of Computational Linguistics, Peking University, MOE, China\\
  $^{2}$School of Software and Microelectronics, Peking University, China\\
  $^{3}$ByteDance AI Lab, China\\
  \texttt{
    \{zengs,chbb\}@pku.edu.cn runxinxu@gmail.com lileilab@bytedance.com
  }
}

\date{}

\begin{document}
\maketitle

\renewcommand{\thefootnote}{\fnsymbol{footnote}} 
\footnotetext[1]{Equal contribution.} 
\footnotetext[2]{Corresponding author.} 
\renewcommand{\thefootnote}{\fnsymbol{footnote}}

\input{tex/abstract.tex}
\input{tex/intro.tex}
\input{tex/task_formulation.tex}

\input{tex/model.tex}
\input{tex/experiments.tex}
\input{tex/related.tex}
\input{tex/conclusion.tex}

\section*{Acknowledgments}
The authors would like to thank the anonymous reviewers for their thoughtful and constructive comments and ByteDance AI Lab for providing the computational resources for this work.
This paper is supported in part by the National Key R\&D Program of China under Grand No.2018AAA0102003, the National Science Foundation of China under Grant No.61876004 and 61936012.

\bibliography{anthology,emnlp2020}
\bibliographystyle{acl_natbib}

\input{tex/appendix.tex}




\end{document}

%% file: tex/abstract.tex
\begin{abstract}

Document-level relation extraction aims to extract relations among entities within a document. 
Different from sentence-level relation extraction, it requires reasoning over multiple sentences across a document. 
In this paper, we propose \textbf{G}raph \textbf{A}ggregation-and-\textbf{I}nference \textbf{N}etwork (GAIN) featuring double graphs. 
GAIN first constructs a \textit{heterogeneous mention-level graph} (hMG) to model complex interaction among different mentions across the document. 
It also constructs an \textit{entity-level graph} (EG), based on which we propose a novel path reasoning mechanism to infer relations between entities. Experiments on the public dataset, DocRED, show GAIN achieves a significant performance improvement ($2.85$ on F1) over the previous state-of-the-art. Our code is available at \url{https://github.com/DreamInvoker/GAIN}.

\end{abstract}

%% file: tex/intro.tex
\section{Introduction}
The task of identifying semantic relations between entities from text, namely relation extraction (RE), plays a crucial role in a variety of knowledge-based applications, such as question answering (\citealp{yu-etal-17-improved}) and large-scale knowledge graph construction.
\begin{figure}
    \centering
    \includegraphics[width=1.0\linewidth]{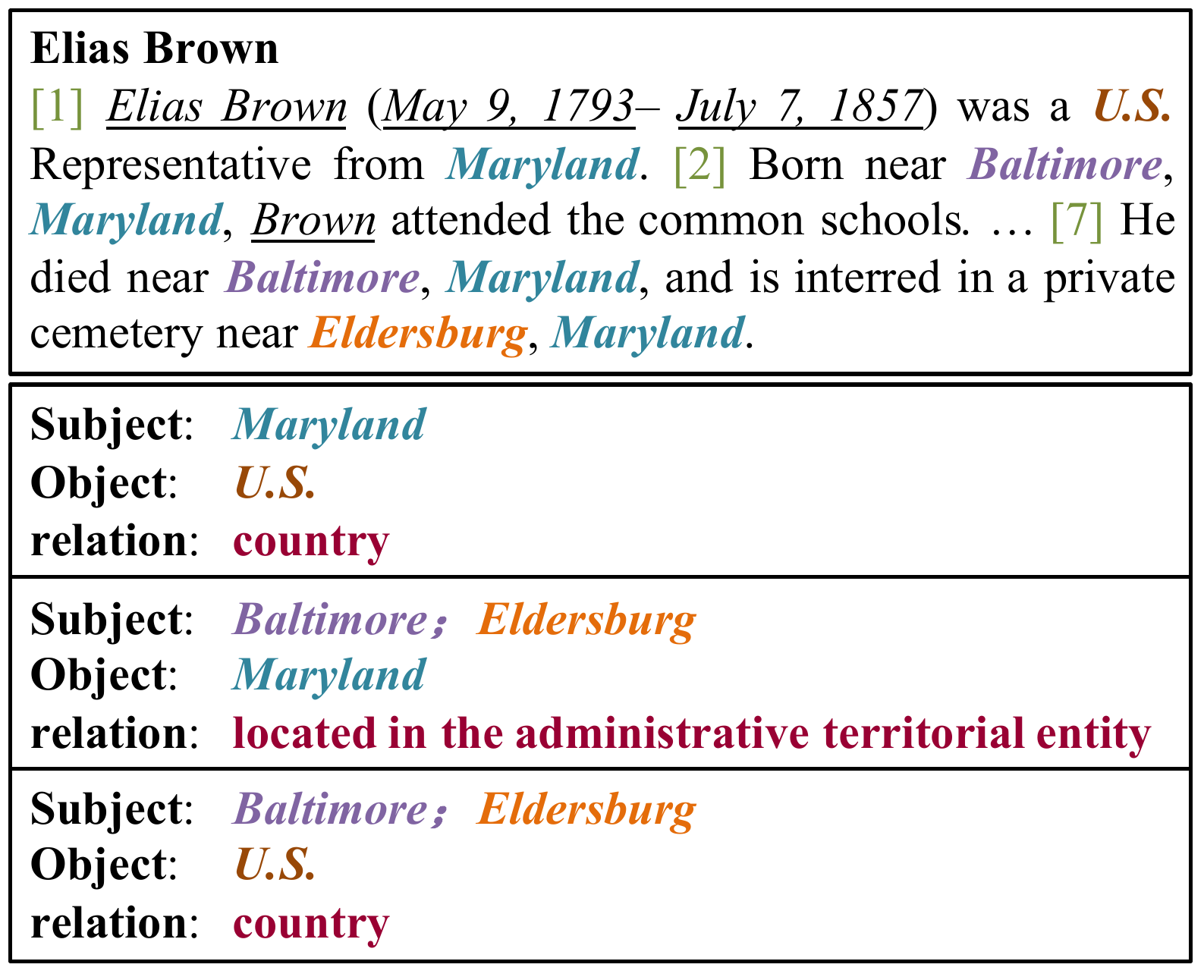}
    \caption{An example document and its desired relations from DocRED \citep{yao-etal-19-docred}. Entity mentions and relations involved in these relation instances are colored. Other mentions are underlined for clarity.}
    \label{fig:running-example}
\end{figure}
Previous methods (\citealp{zeng-etal-14-relation}; \citealp{zeng-etal-15-distant}; \citealp{xiao-liu-16-semantic}; \citealp{zhang-etal-17-position}; \citealp{zhang-etal-18-graph}; \citealp{baldini-soares-etal-19-matching}) focus on sentence-level RE, which predicts relations among entities in a single sentence.
However, sentence-level RE models suffer from an inevitable limitation -- they fail to recognize relations between entities across sentences. 
Hence, extracting relations at the document-level is necessary for a holistic understanding of knowledge in text.

There are several major challenges in effective relation extraction at the document-level.
Firstly, the subject and object entities involved in a relation may appear in different sentences. Therefore a relation cannot be identified based solely on a single sentence.
Secondly, the same entity may be mentioned multiple times in different sentences. Cross-sentence context information has to be aggregated to represent the entity better.
Thirdly, the identification of many relations requires techniques of logical reasoning. This means these relations can only be successfully extracted when other entities and relations, usually spread across sentences, are identified implicitly or explicitly.  
As Figure~\ref{fig:running-example} shows, it is easy to recognize the intra-sentence relations (\textit{Maryland}, country, \textit{U.S.}), (\textit{Baltimore}, located in the administrative territorial entity, \textit{Maryland}), and (\textit{Eldersburg}, located in the administrative territorial entity, \textit{Maryland}), since the subject and object appear in the same sentence.
However, it is non-trivial to predict the inter-sentence relations between \textit{Baltimore} and \textit{U.S.}, as well as \textit{Eldersburg} and \textit{U.S.}, whose mentions do not appear in the same sentence and have long-distance dependencies.
Besides, the identification of these two relation instances also requires logical reasoning. For example, \textit{Eldersburg} belongs to \textit{U.S.} because \textit{Eldersburg} is located in \textit{Maryland}, which belongs to \textit{U.S.}.

Recently, \citet{yao-etal-19-docred} proposed a large-scale human-annotated document-level RE dataset, DocRED, to push sentence-level RE forward to document-level and it contains massive relation facts. Figure~\ref{fig:running-example} shows an example from DocRED. 
We randomly sample 100 documents from the DocRED dev set and manually analyze the bad cases predicted by a BiLSTM-based model proposed by \citet{yao-etal-19-docred}. As shown in Table~\ref{table:bad-cases}, the error type of inter-sentence and that of logical reasoning take up a large proportion of all bad cases, with $53.5\%$ and $21.0\%$ respectively.
Therefore, in this paper, we aim to tackle these problems to extract relations from documents better.

\begin{table}
\centering
\begin{tabular}{lc}
\hline
Error Type & Count \\
\hline
Intra-sentence & 535 \\
Inter-sentence & 615 \\
\hline
\hline
Logical Reasoning & 242 \\
\hline
\end{tabular}
\caption{Statistics of bad cases in randomly sampled 100 documents from DocRED dev set for BiLSTM \citep{yao-etal-19-docred}, with 1150 bad cases in total.}
\label{table:bad-cases}
\end{table}

Previous work in document-level RE do not consider reasoning (\citealp{DBLP:conf/aaai/GuptaRSR19}; \citealp{jia-etal-19-document}; \citealp{yao-etal-19-docred}), or only use graph-based or hierarchical neural network to conduct reasoning in an implicit way (\citealp{peng-etal-17-cross}; \citealp{sahu-etal-19-inter}; \citealp{LSR}).
In this paper, we propose a \textbf{G}raph \textbf{A}ggregation-and-\textbf{I}nference \textbf{N}etwork (GAIN) for document-level relation extraction. 
It is designed to tackle the challenges mentioned above directly.
GAIN constructs a \textbf{h}eterogeneous \textbf{M}ention-level \textbf{G}raph (hMG) with two types of nodes, namely mention node and document node, and three different types of edges, i.e., intra-entity edge, inter-entity edge and document edge, to capture the context information of entities in the document.
Then, we apply Graph Convolutional Network \citep{kipf2017semi} on hMG to get a document-aware representation for each mention. 
\textbf{E}ntity-level \textbf{G}raph (EG) is then constructed by merging mentions that refer to the same entity in hMG, on top of which we propose a novel path reasoning mechanism. This reasoning mechanism allows our model to infer multi-hop relations between entities.

In summary, our main contributions are as follows:
\begin{itemize}
    \item We propose a novel method, Graph Aggregation-and-Inference Network (GAIN), which features a double graph design, to better cope with document-level RE task.
    
    \item We introduce a heterogeneous Mention-level Graph (hMG) with a graph-based neural network to model the interaction among different mentions across the document and offer document-aware mention representations.
    
    \item We introduce an Entity-level Graph (EG) and propose a novel path reasoning mechanism for relational reasoning among entities.
\end{itemize}

We evaluate GAIN on the public DocRED dataset. It significantly outperforms the previous state-of-the-art model by $2.85$ F1 score. 
Further analysis demonstrates the capability of GAIN to aggregate document-aware context information and to infer logical relations over documents. 

%% file: tex/task_formulation.tex
\section{Task Formulation}
We formulate the document-level relation extraction task as follows. Given a document comprised of $N$ sentences $\mathcal{D}=\left\{s_i\right\}^{N}_{i=1}$ and a variety of entities $\mathcal{E}=\left\{e_i\right\}^{P}_{i=1}$, where $s_i=\left\{w_j\right\}^{M}_{j=1}$ refers to the $i$-th sentence consisting of $M$ words, $e_i=\left\{m_j\right\}^{Q}_{j=1}$ and $m_j$ refers to a span of words belonging to the $j$-th mention of the $i$-th entity, the task aims to extract the relations between different entities in $\mathcal{E}$, namely $\left\{(e_i, r_{ij}, e_j) | e_i, e_j \in \mathcal{E}, r_{ij} \in \mathcal{R} \right\}$, where $\mathcal{R}$ is a pre-defined relation type set.

In our paper, a relation $r_{ij}$ between entity $e_i$ and $e_j$ is defined as inter-sentential, if and only if $S_{e_i}\cap S_{e_j} = \varnothing $, where $S_{e_i}$ denotes those sentences containing mentions of $e_i$. 
Instead, a relation $r_{ij}$ is defined as intra-sentential, if and only if $S_{e_i}\cap S_{e_j} \neq \varnothing $.
We also define $K$-hop relational reasoning as predicting relation $r_{ij}$ based on a $K$-length chain of existing relations, with $e_i$ and $e_j$ being the head and tail of the reasoning chain, i.e., $e_i \stackrel{r_1}{\longrightarrow} e_{m} \stackrel{r_2}{\longrightarrow} \dots e_n \stackrel{r_K}{\longrightarrow} e_j \Rightarrow e_i \stackrel{r_{ij}}{\longrightarrow} e_{j} $.




%% file: tex/model.tex
\begin{figure*}
    \centering
    \includegraphics[width=1.0\linewidth]{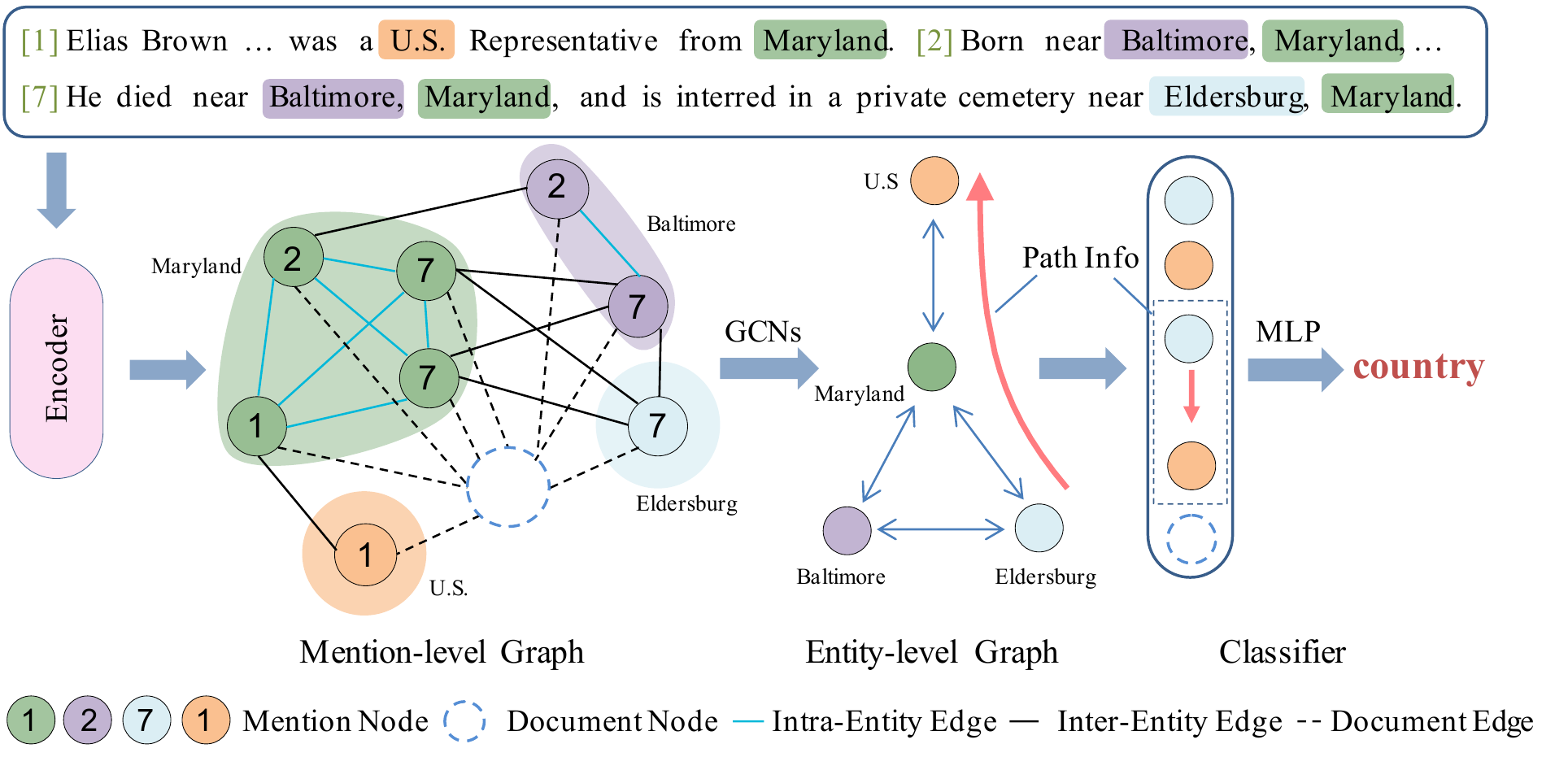}
    \caption{The overall architecture of GAIN. First, A context encoder consumes the input document to get a contextualized representation of each word. Then, the Mention-level Graph is constructed with mention nodes and a document node. After applying GCN, the graph is transformed into Entity-level Graph, where the paths between entities are identified for reasoning. Finally, the classification module predicts target relations based on the above information. Different entities are in different colors. The number $i$ in the mention node denotes that it belongs to the $i$-th sentence.}
    \label{fig:model}
\end{figure*}
\section{Graph Aggregation and Inference Network (GAIN)\label{sec:model}}

GAIN mainly consists of 4 modules: 
encoding module (Sec.~\ref{ssec:encoding}), 
mention-level graph aggregation module (Sec.~\ref{ssec:aggregation}),
entity-level graph inference module (Sec.~\ref{ssec:inference}), classification module (Sec.~\ref{ssec:classification}), as is shown in Figure~\ref{fig:model}.

\subsection{Encoding Module\label{ssec:encoding}}
In the encoding module, we convert a document $\mathcal{D}= \left\{w_i\right\}^{n}_{i=1}$ containing $n$ words into a sequence of vectors $\left\{g_{i}\right\}^{n}_{i=1}$.
Following \citet{yao-etal-19-docred}, for each word $w_i$ in $\mathcal{D}$, we first concatenate its word embedding with entity type embedding and coreference embedding:
\begin{equation}
    x_{i} = [E_w(w_i); E_t(t_i); E_c(c_i)]
\end{equation}
where $E_w(\cdot)$ , $E_t(\cdot)$ and $E_c(\cdot)$ denote the word embedding layer, entity type embedding layer and coreference embedding layer, respectively. $t_i$ and $c_i$ are named entity type and entity id. 
We introduce \textit{None} entity type and id for those words not belonging to any entity.

Then the vectorized word representations are fed into an encoder to obtain the context sensitive representation for each word:
\begin{equation}
    [g_{1}, g_{2}, \ldots, g_{n}] = Encoder([x_1, x_2, \ldots, x_n])
\end{equation}
where the $Encoder$ can be LSTM or other models.

\subsection{Mention-level Graph Aggregation Module\label{ssec:aggregation}}
To model the document-level information and interactions between mentions and entities, a heterogeneous Mention-level Graph (hMG) is constructed.

hMG has two different kinds of nodes: mention node and document node. Each mention node denotes one particular mention of an entity.
And hMG also has one document node that aims to model the overall document information. We argue that this node could serve as a pivot to interact with different mentions and thus reduce the long distance among them in the document.

There are three types of edges in hMG:
\begin{itemize}
    \item \textbf{Intra-Entity Edge:} Mentions referring to the same entity are fully connected with intra-entity edges. In this way, the interaction among different mentions of the same entity could be modeled.
 
    \item \textbf{Inter-Entity Edge:} Two mentions of different entities are connected with an inter-entity edge if they co-occur in a single sentence. In this way, interactions among entities could be modeled by co-occurrences of their mentions. 
    \item \textbf{Document Edge:} All mentions are connected to the document node with the document edge. With such connections, the document node can attend to all the mentions and enable interactions between document and mentions. Besides, the distance between two mention nodes is at most two with the document node as a pivot. Therefore long-distance dependency can be better modeled.
\end{itemize}


Next, we apply Graph Convolution Network \citep{kipf2017semi} on hMG to aggregate the features from neighbors. Given node $u$ at the $l$-th layer, the graph convolutional operation can be defined as:
\begin{equation}
       h_{u}^{(l + 1)} = \sigma \left(\sum_{k\in\mathcal{K}}\sum_{v\in\mathcal{N}_k(u)} W^{(l)}_k h_{v}^{(l)} + b^{(l)}_k\right)
\end{equation}
where $\mathcal{K}$ are different types of edges, $W^{(l)}_k\in \mathbb{R}^{d\times d}$ and $b^{(l)}_k \in \mathbb{R}^{d}$ are trainable parameters.
$\mathcal{N}_k(u)$ denotes neighbors for node $u$ connected in $k$-th type edge.
$\sigma$ is an activation function (e.g., ReLU).

Different layers of GCN express features of different abstract levels, and therefore in order to cover features of all levels, we concatenate hidden states of each layer to form the final representation of node $u$:
\begin{equation}
    \textbf{m}_u = [h_{u}^{(0)}; h_{u}^{(1)}; \ldots; h_{u}^{(N)}]
\end{equation}
where $h^{(0)}_u$ is the initial representation of node $u$. For a mention ranging from the $s$-th word to the $t$-th word in the document, $h^{(0)}_u = \frac{1}{t-s+1} \sum_{j=s}^{t} g_{j}$ and for document node, it is initialized with the document representation output from the encoding module.


\subsection{Entity-level Graph Inference Module\label{ssec:inference}}

In this subsection, we introduce Entity-level Graph (EG) and path reasoning mechanism. 
First, mentions that refer to the same entity are merged to entity node so as to get the nodes in EG. Note that we do not consider document node in EG. For $i$-th entity node $\textbf{e}_i$ mentioned $N$ times, it is represented by the average of its $N$ mention representations:
\begin{equation}
    \textbf{e}_{i} = \frac{1}{N} \sum_{n} \textbf{m}_{n}
\label{entity-rep}
\end{equation}

Then, we merge all inter-entity edges that connect mentions of the same two entities so as to get the edges in EG. The representation of directed edge from $\textbf{e}_i$ to $\textbf{e}_j$ in the EG is defined as :
\begin{equation}
    \textbf{e}_{ij} = \sigma \left(W_{q}[\textbf{e}_{i}; \textbf{e}_{j}] + b_{q}\right)
\end{equation}
where $W_{q}$ and $b_{q}$ are trainable parameters, and $\sigma$ is an activation function (e.g., ReLU).

Based on the vectorized edge representation, the $i$-th path between head entity $\textbf{e}_{h}$ and tail entity $\textbf{e}_{t}$ passing through entity $\textbf{e}_{o}$ is represented as:
\begin{equation}
    \textbf{p}_{h,t}^{i} = [\textbf{e}_{ho};\textbf{e}_{ot};\textbf{e}_{to};\textbf{e}_{oh}]
\end{equation}
Note that we only consider two-hop paths here, while it can easily extend to multi-hop paths.

We also introduce attention mechanism \citep{attention-15}, using the entity pair $\left(\textbf{e}_{h},\textbf{e}_{t}\right)$ as query, to fuse the information of different paths between $\textbf{e}_{h}$ and $\textbf{e}_{t}$. 
\begin{equation}
    s_i = \sigma([\textbf{e}_h; \textbf{e}_t] \cdot W_l \cdot \textbf{p}_{h,t}^i)
\end{equation}
\begin{equation}
    \alpha_i = \frac{e^{s_i}}{\sum_{j}e^{s_j}} 
\end{equation}
\begin{equation}
    \textbf{p}_{h,t} = \sum_{i} \alpha_{i} \textbf{p}_{h,t}^{i}
\end{equation}
where $\alpha_{i}$ is the normalized attention weight for $i$-th path. Consequently, the model will pay more attention to useful paths. $\sigma$ is an activation function.

With this module, an entity can be represented by fusing information from its mentions, which usually spread in multiple sentences. Moreover, potential reasoning clues are modeled by different paths between entities. Then they can be integrated with the attention mechanism so that we will take into account latent logical reasoning chains to predict relations.


\subsection{Classification Module\label{ssec:classification}}
For each entity pair $\left(\textbf{e}_{h}, \textbf{e}_{t}\right)$, we concatenate the following representations: (1) the head and tail entity representation $\textbf{e}_{h}$ and $\textbf{e}_{t}$ derived in the Entity-level Graph, with the comparing operation \citep{mou-etal-16-natural} to strengthen features, i.e., absolute value of subtraction between the representation of two entities, $|\textbf{e}_{h} - \textbf{e}_{t}|$, and element-wise multiplication, $\textbf{e}_{h} \odot \textbf{e}_{t}$; (2) the representation of document node in Mention-level Graph, $\textbf{m}_{doc}$, as it can help aggregate cross-sentence information and provide document-aware representation; (3) the comprehensive inferential path information $\textbf{p}_{h,t}$.
\begin{equation}
    I_{h,t} = [\textbf{e}_{h}; \textbf{e}_{t};  |\textbf{e}_{h} - \textbf{e}_{t}|; \textbf{e}_{h} \odot \textbf{e}_{t};
    \textbf{m}_{doc};\textbf{p}_{h,t} ]
\end{equation}

Finally, we formulate the task as multi-label classification task and predict relations between entities:
\begin{equation}
    P(r|\textbf{e}_{h}, \textbf{e}_{t}) = sigmoid \left(W_{b} \sigma (W_{a}I_{h,t} + b_{a}) + b_{b}\right)
\end{equation}
where $W_{a}$, $W_{b}$, $b_{a}$, $b_{b}$ are trainable parameters, $\sigma$ is an activation function (e.g., ReLU). We use binary cross entropy as the classification loss to train our model in an end-to-end way:
\begin{equation}
\begin{split}
    \mathcal{L} &= - \sum _{\mathcal{D} \in \mathcal{S}}  \sum _{h\neq t} \sum_{r_i \in \mathcal{R}}  \mathbb I \left ( r_i=1 \right ) \log P \left ( r_i|\textbf{e}_{h}, \textbf{e}_{t} \right )
    \\
     & +  \mathbb I \left ( r_i=0 \right ) \log \left(  1 - P \left ( r_i|\textbf{e}_{h}, \textbf{e}_{t} \right ) \right )
\end{split}
\end{equation}
where $\mathcal{S}$ denotes the whole corpus, and $\mathbb I\left(\cdot\right)$ refers to indication function.

%% file: tex/experiments.tex
\section{Experiments\label{sec:experiments}}

\subsection{Dataset}
We evaluate our model on DocRED \citep{yao-etal-19-docred}, a large-scale human-annotated dataset for document-level RE constructed from Wikipedia and Wikidata. 
DocRED has $96$ relations types, $132,275$ entities, and $56,354$ relational facts in total.
Documents in DocRED contain about $8$ sentences on average, and more than $40.7\%$ relation facts can only be extracted from multiple sentences.
Moreover, $61.1\%$ relation instances require various inference skills such as logical inference \citep{yao-etal-19-docred}.
we follow the standard split of the dataset, $3,053$ documents for training, $1,000$ for development and $1,000$ for test. For more detailed statistics about DocRED, we recommend readers to refer to the original paper \citep{yao-etal-19-docred}.

\subsection{Experimental Settings}
In our GAIN implementation, we use $2$ layers of GCN and set the dropout rate to $0.6$, learning rate to $0.001$.
We train GAIN using AdamW \citep{adamW} as optimizer with weight decay $0.0001$ and implement GAIN under PyTorch \citep{paszke2017automatic} and DGL \citep{wang2019dgl}.

We implement three settings for our GAIN. \textbf{GAIN-GloVe} uses GloVe ($100$d) and BiLSTM ($256$d) as word embedding and encoder.
\textbf{GAIN-BERT$_{base}$} and \textbf{GAIN-BERT$_{large}$} use BERT$_{base}$ and BERT$_{large}$ as encoder respectively and the learning rate is set to $1e^{-5}$.

\subsection{Baselines and Evaluation Metrics}

We use the following models as baselines.

\citet{yao-etal-19-docred} proposed models to encode the document into a sequence of hidden state vector $\{\bm{h_i}\}_{i=1}^{n}$ using \textbf{CNN} \citep{cnn}, \textbf{LSTM} \citep{HochreiterS97}, and \textbf{BiLSTM} \citep{650093} as their encoder, and predict relations between entities with their representations.
Other pre-trained models like \textbf{BERT} \citep{DevlinCLT19}, \textbf{RoBERTa} \citep{RoBERTa}, and \textbf{CorefBERT} \citep{CorefBERT} are also used as encoder \citep{wang2019fine, CorefBERT}  to document-level RE task.

\textbf{Context-Aware}, also proposed by \citet{yao-etal-19-docred} on DocRED adapted from \citep{SorokinG17}, uses an LSTM to encode the text, but further utilizes attention mechanism to absorb the context relational information for predicting.


\textbf{BERT-Two-Step$_{base}$}, proposed by \citet{wang2019fine} on DocRED. Though similar to BERT-RE$_{base}$, it first predicts whether two entities have a relationship and then predicts the specific target relation.

\textbf{HIN-GloVe/HIN-BERT$_{base}$}, proposed by \citet{tang2020hin}. Hierarchical Inference Network (HIN) aggregate information from entity-level, sentence-level, and document-level to predict target relations, and use GloVe \citep{PenningtonSM14} or BERT$_{base}$ for word embedding.

\textbf{LSR-GloVe/LSR-BERT$_{base}$}, proposed by \citet{LSR} recently. They construct a graph based on the dependency tree and predict relations by latent structure induction and GCN. \citet{LSR} also adapted four graph-based state-of-the-art RE models to DocRED, including \textbf{GAT} \citep{GAT}, \textbf{GCNN} \citep{sahu-etal-19-inter}, \textbf{EoG} \citep{christopoulou-etal-2019-connecting}, and \textbf{AGGCN} \citep{AGGCN}. We also include their results.

Following \citet{yao-etal-19-docred}, we use the widely used metrics F1 and AUC in our experiment.
We also use Ign F1 and Ign AUC, which calculate F1 and AUC excluding the common relation facts in the training and dev/test sets.

\subsection{Results\label{ssec:quantitative}}
We show GAIN's performance on the DocRED dataset in Table~\ref{table:results}, in comparison with other baselines.

\begin{table*}[htbp]
\centering
\scriptsize
\begin{tabular}{lcccccc}
\hline
\multirow{2}{*}{\bf Model} & \multicolumn{4}{c}{\bf Dev} & \multicolumn{2}{c}{\bf Test}  \\ 
\cmidrule(lr){2-5} \cmidrule(lr){6-7}
~ & \tiny \bf Ign F1 & \tiny \bf Ign AUC & \tiny \bf F1 & \tiny \bf AUC & \tiny \bf Ign F1 & \tiny \bf F1 \\
\hline
CNN$^*$ \citep{yao-etal-19-docred} & 41.58 & 36.85 & 43.45 & 39.39 & 40.33 & 42.26 \\
LSTM$^*$ \citep{yao-etal-19-docred} & 48.44 & 46.62 & 50.68 & 49.48 & 47.71 & 50.07 \\
BiLSTM$^*$ \citep{yao-etal-19-docred} & 48.87 & 47.61 & 50.94 & 50.26 & 48.78 & 51.06 \\
Context-Aware$^*$ \citep{yao-etal-19-docred} & 48.94 & 47.22 & 51.09 & 50.17 & 48.40 & 50.70 \\
HIN-GloVe$^*$ \citep{tang2020hin} & 51.06 & - & 52.95 & - & 51.15 & 53.30 \\
\hline
GAT$^{\ddagger}$ \citep{GAT} & 45.17 & - & 51.44 & - & 47.36 & 49.51 \\
GCNN$^{\ddagger}$ \citep{sahu-etal-19-inter} & 46.22 & - & 51.52 & - & 49.59 & 51.62 \\
EoG$^{\ddagger}$ \citep{christopoulou-etal-2019-connecting} & 45.94 & - & 52.15 & - & 49.48 & 51.82 \\
AGGCN$^{\ddagger}$ \citep{AGGCN} & 46.29 & - & 52.47 & - & 48.89 & 51.45 \\
LSR-GloVe$^*$ \citep{LSR} & 48.82 & - & 55.17 & - & 52.15 & 54.18 \\
\hline
GAIN-GloVe & \textbf{53.05} & \textbf{52.57} & \textbf{55.29} & \textbf{55.44} & \textbf{52.66} & \textbf{55.08} \\
\hline 
\hline
BERT-RE$_{base}^*$ \citep{wang2019fine} & - & - & 54.16 & - & - & 53.20 \\
RoBERTa-RE$_{base}^\dagger $  & 53.85 & 48.27 & 56.05 & 51.35 & 53.52 & 55.77 \\
BERT-Two-Step$_{base}^*$ \citep{wang2019fine} & - & - & 54.42 & - & - & 53.92 \\
HIN-BERT$_{base}^*$ \citep{tang2020hin} & 54.29 & - & 56.31 & - & 53.70 & 55.60 \\
CorefBERT-RE$_{base}^*$ \citep{CorefBERT} & 55.32 & - & 57.51 & - & 54.54 & 56.96 \\
LSR-BERT$_{base}^*$ \citep{LSR} & 52.43 & - & 59.00 & - & 56.97 & 59.05 \\
\hline
GAIN-BERT$_{base}$ & \textbf{59.14} & \textbf{57.76} & \textbf{61.22} & \textbf{60.96} & \textbf{59.00} & \textbf{61.24} \\
\hline
\hline
BERT-RE$_{large}^*$ \citep{CorefBERT} & 56.67 & - & 58.83 & - & 56.47 & 58.69\\
CorefBERT-RE$_{large}^*$ \citep{CorefBERT} & 56.73 & - & 58.88 & - & 56.48 & 58.70\\
RoBERTa-RE$_{large}^*$ \citep{CorefBERT} & 57.14 & - & 59.22 & - & 57.51 & 59.62\\
CorefRoBERTa-RE$_{large}^*$ \citep{CorefBERT} & 57.84 & - & 59.93 & - & 57.68 & 59.91 \\
\hline
GAIN-BERT$_{large} $ & \textbf{60.87} & \textbf{61.79} & \textbf{63.09} & \textbf{64.75} & \textbf{60.31} & \textbf{62.76} \\
\hline
\end{tabular}
\caption{Performance on DocRED. Models above the first double line do not use pre-trained model. Results with * are reported in their original papers. Results with $\ddagger$ are performances of graph-based state-of-the-art RE models implemented in \citep{LSR}. Results with $\dagger$ are based on our implementation.
}
\label{table:results}
\end{table*}

Among the models not using BERT or BERT variants, GAIN-GloVe consistently outperforms all sequential-based and graph-based strong baselines by $0.9\sim12.82$ F1 score on the test set. 
Among the models using BERT or BERT variants, GAIN-BERT$_{base}$ yields a great improvement of F1/Ign F1 on dev and test set by $2.22/6.71$ and $2.19/2.03$, respectively, in comparison with the strong baseline LSR-BERT$_{base}$. GAIN-BERT$_{large}$ also improves $2.85$/$2.63$ F1/Ign F1 on test set compared with previous state-of-the-art method, CorefRoBERTa-RE$_{large}$. It suggests that GAIN is more effective in document-level RE tasks.
We can also observe that LSR-BERT$_{base}$ improves F1 by $3.83$ and $4.87$ on dev and test set with GloVe embedding replaced with BERT$_{base}$. In comparison, our GAIN-BERT$_{base}$ yields an improvement by $5.93$ and $6.16$, which indicates GAIN can better utilize BERT representation.


\subsection{Ablation Study\label{ssec:ablation}}

To further analyze GAIN, we also conduct ablation studies to illustrate the effectiveness of different modules and mechanisms in GAIN. 
We show the results of the ablation study in Table~\ref{table:ablation}.

\begin{table*}
\centering
\begin{tabular}{lcccccc}
\hline
\multirow{2}{*}{Model} & \multicolumn{4}{c}{Dev} & \multicolumn{2}{c}{Test}  \\ 
\cmidrule(lr){2-5} \cmidrule(lr){6-7}
~ & Ign F1 & Ign AUC & F1 & AUC & Ign F1 & F1 \\
\hline
GAIN-GloVe & \textbf{53.05} & \textbf{52.57} & \textbf{55.29} & \textbf{55.44} & \textbf{52.66} & \textbf{55.08} \\
\quad \textit{- hMG} & 50.97 & 48.84 & 53.10 & 51.73 & 50.76 & 53.06 \\
\quad \textit{- Inference Module} & 50.84 & 48.68 & 53.02 & 51.58 & 50.32 & 52.66 \\
\quad \textit{- Document Node} & 50.86 & 48.68 & 53.01 & 52.46 & 50.32 & 52.67 \\
\hline
GAIN-BERT$_{base}$ & \textbf{59.14} & \textbf{57.76} & \textbf{61.22} & \textbf{60.96} & \textbf{59.00} & \textbf{61.24} \\
\quad \textit{- hMG} & 57.12 & 51.54 & 59.17 & 54.61 & 57.31 & 59.56 \\
\quad \textit{- Inference Module} & 56.97 & 54.29 & 59.28 & 57.25 & 57.01 & 59.34 \\
\quad \textit{- Document Node} & 57.26 & 52.07 & 59.62 & 55.51 & 57.01 & 59.63 \\
\hline
\end{tabular}
\caption{Performance of GAIN with different embeddings and submodules.}
\label{table:ablation}
\end{table*}

First, we remove the heterogeneous Mention-level Graph (hMG) of GAIN. In detail, we initialize an entity node in Entity-level Graph (EG) with Eq.~\ref{entity-rep} but replace $\textbf{m}_n$ with $h_n^{(0)}$, and apply GCN to EG instead. Features in different layers of GCN are concatenated to obtain $\textbf{e}_i$.
Without hMG, the performance of GAIN-GloVe/GAIN-BERT$_{base}$ sharply drops by $2.08$/$2.02$ Ign F1 score on dev set. This drop shows that hMG plays a vital role in capturing interactions among mentions belonging to the same and different entities and document-aware features.

Next, we remove the inference module. To be specific, the model abandon the path information between head and tail entity $\textbf{p}_{h,t}$ obtained in Entity-level Graph, and predict relations only based on entity representation, $\textbf{e}_h$ and $\textbf{e}_t$, and document node representation, $\textbf{m}_{doc}$. The inference module's removal results in poor performance across all metrics, for instance,  $2.21$/$2.17$ Ign F1 score decrease on the dev set for GAIN-GloVe/GAIN-BERT$_{base}$. 
It suggests that our path inference mechanism helps capture the potential $K$-hop inference paths to infer relations and, therefore, improve document-level RE performance.

Moreover, taking away the document node in hMG leads to $2.19$/$1.88$ Ign F1 decrease on the dev set for GAIN-GloVe/GAIN-BERT$_{base}$.
It helps GAIN aggregate the document information and works as a pivot to facilitate the information exchange among different mentions, especially those far away from each other within the document.

\begin{table}
\centering
\begin{tabular}{lcc}
\hline
Model & Intra-F1 & Inter-F1 \\
\hline
CNN$^*$ & 51.87 & 37.58 \\
LSTM$^*$ & 56.57 & 41.47 \\
BiLSTM$^*$ & 57.05 & 43.49\\
Context-Aware$^*$ & 56.74 & 42.26\\
LSR-GloVe$^*$ & 60.83 & 48.35 \\
\hline
GAIN-GloVe & \textbf{61.67} & \textbf{48.77} \\
\textit{- hMG} & 59.72 & 46.49 \\
\hline 
\hline
BERT-RE$_{base}^*$ & 61.61 & 47.15\\
RoBERTa-RE$_{base}$ & 65.65 & 50.09 \\
BERT-Two-Step$_{base}^*$ & 61.80 & 47.28 \\
LSR-BERT$_{base}^*$ & 65.26 & 52.05 \\
\hline
GAIN-BERT$_{base}$ & \textbf{67.10} & \textbf{53.90}\\
\textit{- hMG} & 66.15 & 51.42\\
\hline
\end{tabular}
\caption{Intra- and Inter-F1 results on dev set of DocRED. Results with * are reported in \citep{LSR}.}
\label{table:inter}
\end{table}

\subsection{Analysis \& Discussion \label{ssec:qualitative}}
In this subsection, we further analyze both inter-sentential and inferential performance on the development set. The same as \citet{LSR}, we report Intra-F1/Inter-F1 scores in Table~\ref{table:inter}, which only consider either intra- or inter-sentence relations respectively. Similarly, in order to evaluate the inference ability of the models, Infer-F1 scores are reported in Table~\ref{table:infer}, which only considers relations that engaged in the relational reasoning process
. 
For example, we take into account the golden relation facts $r_1$, $r_2$, and $r_3$ if there exist $e_h \stackrel{r_1}{\longrightarrow} e_{o} \stackrel{r_2}{\longrightarrow} e_{t}$ and $e_h \stackrel{r_3}{\longrightarrow} e_{t}$ when calculating Infer-F1.

As Table~\ref{table:inter} shows, GAIN outperforms other baselines not only in Intra-F1 but also Inter-F1, and the removal of hMG leads to a more considerable decrease in Inter-F1 than Intra-F1, which indicates our hMG do help interactions among mentions, especially those distributed in different sentences with long-distance dependency. 

Besides, Table~\ref{table:infer} suggests GAIN can better handle relational inference. For example, GAIN-BERT$_{base}$ improves $5.11$ Infer-F1 compared with 
RoBERTa-RE$_{base}$. The inference module also plays an important role in capturing potential inference chains between entities, without which GAIN-BERT$_{base}$ would drop by $1.78$ Infer-F1.

\begin{table}
\centering
\begin{tabular}{lccc}
\hline
Model & Infer-F1 & P & R\\
\hline
CNN & 37.11 & 32.81 & 42.72\\
LSTM & 39.03 & 33.16 & 47.44\\
BiLSTM & 38.73 & 31.60 & 50.01\\
Context-Aware & 39.73 & \textbf{33.97} & 47.85\\
\hline
GAIN-GloVe & \textbf{40.82}  & 32.76 & \textbf{54.14}\\
\textit{- Inference Module} & 39.76 & 32.26 & 51.80\\
\hline 
\hline
BERT-RE$_{base}$ & 39.62 & 34.12 & 47.23 \\
RoBERTa-RE$_{base}$ & 41.78 & 37.97 & 46.45 \\
\hline
GAIN-BERT$_{base}$ & \textbf{46.89} & \textbf{38.71} & \textbf{59.45}\\
\textit{- Inference Module} & 45.11 & 36.91 & 57.99\\
\hline
\end{tabular}
\caption{Infer-F1 results on dev set of DocRED. P: Precision, R: Recall.}
\label{table:infer}
\end{table}

\begin{figure*}
    \centering
    \includegraphics[width=1.0\linewidth]{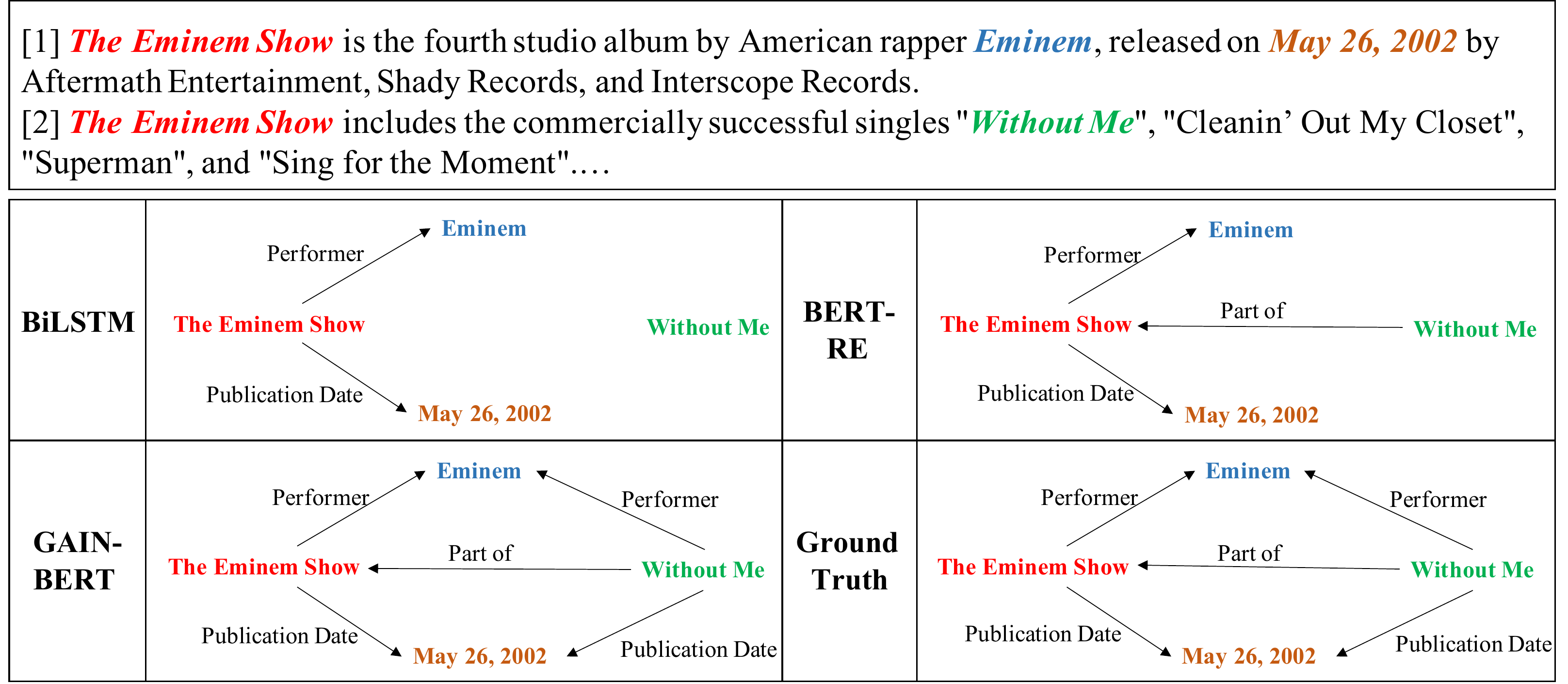}
    \caption{The case study of our proposed GAIN and baseline models. The models take the document as input and predict relations among different entities in different colors. We only show a part of entities within the documents and the according sentences due to the space limitation.}
    \label{fig:experiment_case}
\end{figure*}
\subsection{Case Study}
Figure~\ref{fig:experiment_case} also shows the case study of our proposed model GAIN, in comparison with other baselines. As is shown, BiLSTM can only identify two relations within the first sentence. 
Both BERT-RE$_{base}$ and GAIN-BERT$_{base}$ can successfully predict \textit{Without Me} is part of \textit{The Eminem Show}. But only GAIN-BERT$_{base}$ is able to deduce the performer and publication date of \textit{Without Me} are the same as those of \textit{The Eminem Show}, namely \textit{Eminem} and \textit{May 26, 2002}, where it requires logical inference across sentences.


%% file: tex/related.tex
\section{Related Work}
Previous approaches focus on sentence-level relation extraction (\citealp{zeng-etal-14-relation};
\citealp{zeng-etal-15-distant};
\citealp{wang-etal-16-relation}; \citealp{zhou-etal-16-attention}; \citealp{xiao-liu-16-semantic};
\citealp{zhang-etal-17-position};
\citealp{feng2018reinforcement}; \citealp{zhu-etal-19-graph}). But sentence-level RE models face an inevitable restriction in practice, where many real-world relation facts can only be extracted across sentences. Therefore, many researchers gradually shift their attention into document-level relation extraction.

Several approaches (\citealp{quirk-poon-17-distant}; \citealp{peng-etal-17-cross}; \citealp{DBLP:conf/aaai/GuptaRSR19}; \citealp{song-etal-2018}; \citealp{jia-etal-19-document}) leverage dependency graph to better capture document-specific features, but they ignore ubiquitous relational inference in document. 
Recently, many models are proposed to address this problem.
\citet{tang2020hin} proposed a hierarchical inference network by considering information from entity-level, sentence-level, and document-level. 
However, it conducts relational inference implicitly based on a hierarchical network while we adopt the path reasoning mechanism, which is a more explicit way.

\citep{christopoulou-etal-2019-connecting} is one of the most powerful systems on document-level RE tasks recently. 
Compared to \citep{christopoulou-etal-2019-connecting} and other graph-based approaches to relation extraction, our architecture features many different designs with different motivations behind them. First, the ways of graph construction are different. We create two separate graphs of different levels to capture long-distance document-aware interactions and entity path inference information, respectively. While \citet{christopoulou-etal-2019-connecting} put mentions and entities in the same graph. Moreover, they do not conduct graph node representation learning like GCN to aggregate interactive information on the constructed graph, only using the features from BiLSTMs to represent nodes. Second, the processes of path inference are different. \citet{christopoulou-etal-2019-connecting} use a walk-based method to iteratively generate a path for every entity pair, which requires the extra overhead of hyper-parameter tuning to control the process of inference. Instead, we use an attention mechanism to selectively fuse all possible path information for the entity pair while without extra overhead.

When we were writing this paper, \citep{LSR} make their work public as preprints, which adopt the dependency tree to capture the semantic information in the document. They put mention and entity nodes in the same graph and conduct inference implicitly by using GCN.
Unlike their work, our GAIN presents mention node and entity node in different graphs to better conduct inter-sentence information aggregation and infer relations more explicitly. 

Some other attempts (\citealp{verga-2018-simultaneously, sahu-etal-19-inter, christopoulou-etal-2019-connecting}) study document-level RE in a specific domain like biomedical RE.
However, the datasets they use usually contain very limited relation types and entity types. 
For instance, CDR \citep{li2016biocreative} only has one type of relation and two types of entities, which may not be the ideal testbed for relational reasoning. 

%% file: tex/conclusion.tex
\section{Conclusion}





Extracting inter-sentence relations and conducting relational reasoning are challenging in document-level relation extraction. 

In this paper, we introduce Graph Aggregation-and-Inference Network (GAIN) to better cope with document-level relation extraction, which features double graphs in different granularity.
GAIN utilizes a heterogeneous Mention-level Graph to model the interaction among different mentions across the document and capture document-aware features.
It also uses an Entity-level Graph with a proposed path reasoning mechanism to infer relations more explicitly.

Experimental results on the large-scale human-annotated dataset, DocRED, show GAIN outperforms previous methods, especially in inter-sentence and inferential relations scenarios. 
The ablation study also confirms the effectiveness of different modules in our model.

%% file: tex/appendix.tex
\appendix

\section{Hyperparameter settings\label{sec:appendix}}
We use development set to manually tune the optimal hyperparameters for GAIN, based on the Ign F1 score. 
Hyperparameter settings for GAIN-GloVe, GAIN-BERT$_{base}$ and GAIN-BERT$_{large}$ are listed in Table~\ref{tab:hyperparam1}, ~\ref{tab:hyperparam2} and ~\ref{tab:hyperparam3}, respectively. The value of hyperparameters we finally adopted are in bold. Note that we do not tune all the hyperparameters.



\begin{table}[!htbp]
\centering
\begin{tabular}{lr}
\hline
\textbf{Hyperparameter} & Value \\
\hline
Batch Size &  16, \textbf{32} \\
Learning Rate  & \textbf{0.001}  \\
Activation Function & \textbf{ReLU}, Tanh \\
Positive v.s. Negative Ratio & 1, 0.5, \textbf{0.25} \\
Word Embedding Size  & \textbf{100}\\
Entity Type Embedding Size & \textbf{20}  \\
Coreference Embedding Size & \textbf{20}  \\
Encoder Hidden Size &  128, \textbf{256} \\
Dropout &  0.2, \textbf{0.6}, 0.8 \\
Layers of GCN & 1, \textbf{2}, 3 \\
GCN Hidden Size & \textbf{512} \\
Weight Decay & \textbf{0.0001} \\
\hline \hline
Numbers of Parameters & 63M \\ 
Hyperparameter Search Trials & 12 \\
\hline
\end{tabular}
\caption{Settings for GAIN-GloVe.}
\label{tab:hyperparam1}
\end{table}

\begin{table}[!htbp]
\centering
\begin{tabular}{lr}
\hline
\textbf{Hyperparameter} & Value \\
\hline
Batch Size & \textbf{5} \\
Learning Rate  & \textbf{0.001}  \\
Activation Function & \textbf{ReLU}, Tanh \\
Positive v.s. Negative Ratio & 1, 0.5, \textbf{0.25} \\
Entity Type Embedding Size & \textbf{20} \\
Coreference Embedding Size & \textbf{20} \\
Dropout & 0.2, \textbf{0.6}, 0.8 \\
Layers of GCN & 1, \textbf{2}, 3 \\
GCN Hidden Size & \textbf{808} \\
Weight Decay & \textbf{0.0001} \\
\hline \hline
Numbers of Parameters & 217M \\ 
Hyperparameter Search Trials & 20 \\
\hline
\end{tabular}
\caption{Settings for GAIN-BERT$_{base}$.}
\label{tab:hyperparam2}
\end{table}

\begin{table}[!htbp]
\centering
\begin{tabular}{lr}
\hline
\textbf{Hyperparameter} & Value\\
\hline
Batch Size & \textbf{5}\\
Learning Rate  & \textbf{0.001} \\
Activation Function &  \textbf{ReLU}, Tanh \\
Positive v.s. Negative Ratio & 1, 0.5, \textbf{0.25} \\
Entity Type Embedding Size & \textbf{20} \\
Coreference Embedding Size & \textbf{20} \\
Dropout & 0.2, \textbf{0.6}, 0.8 \\
Layers of GCN & 1, \textbf{2}, 3 \\
GCN Hidden Size & \textbf{1064} \\
Weight Decay & \textbf{0.0001} \\
\hline \hline
Numbers of Parameters &  512M \\ 
Hyperparameter Search Trials & 20 \\
\hline
\end{tabular}
\caption{Settings for GAIN-BERT$_{large}$.}
\label{tab:hyperparam3}
\end{table}